\documentclass[twoside,11pt]{article}

\usepackage{jmlr2e}
\usepackage{subfig}
\usepackage{paralist}
\usepackage{todonotes}
\usepackage{bm}
\newcommand{\cave}{\textit{CAVE}}
\newcommand{\dc}{\textsc{DeepCAVE}}

\usepackage{setspace} 
\usepackage{etoolbox}
\AtBeginEnvironment{quote}{\par\singlespacing\small}

\jmlrheading{1}{2021}{1-48}{4/00}{10/00}{Sass, Bergman, Biedenkapp, Hutter and Lindauer}

\ShortHeadings{DeepCAVE}{Sass, Bergman, Biedenkapp, Hutter and Lindauer}
\firstpageno{1}

\begin{document}

\title{DeepCAVE: An Interactive Analysis Tool for\\ Automated Machine Learning}

\author{\name René Sass$^1$ \email sass@tnt.uni-hannover.de \\
        \name Eddie Bergman$^2$ \email bergmane@cs.uni-freiburg.de\\
        \name André Biedenkapp$^2$\email biedenka@cs.uni-freiburg.de\\
        \name Frank Hutter$^{2,3}$\email fh@cs.uni-freiburg.de\\
        \name Marius Lindauer$^1$ \email lindauer@tnt.uni-hannover.de\\
       \addr $^1$Leibniz University Hannover,
       \addr $^2$University of Freiburg,
       \addr $^3$Bosch Center for Artificial Intelligence
}

\editor{TBA}

\maketitle

\begin{abstract}
    Automated Machine Learning (AutoML) is used more than ever before to support users in determining efficient hyperparameters, neural architectures, or even full machine learning pipelines.
    However, users tend to mistrust the optimization process and its results due to a lack of transparency, making manual tuning still widespread.
    We introduce \dc, an interactive framework to analyze and monitor state-of-the-art optimization procedures for AutoML easily and ad hoc.
    By aiming for full and accessible transparency, \dc{} builds a bridge between users and AutoML and contributes to establishing trust. 
    Our framework's modular and easy-to-extend nature provides users with automatically generated text, tables, and graphic visualizations.
    We show the value of \dc{} in an exemplary use-case of outlier detection, in which our framework makes it easy to identify problems, compare multiple runs and interpret optimization processes.
    The package is freely available on GitHub \url{https://github.com/automl/DeepCAVE}.
\end{abstract}

\section{Introduction}

Experimental design can be tedious and error-prone in practice, particularly if there is little feasible insight into the metric to be optimized and high-dimensional problems with many (hyper-)parameters are tackled.
A typical scenario of experimental design is the tuning of machine learning systems, including data processing, various architectures of deep neural networks, and several hyperparameters for all components in the pipeline.
Automated Machine Learning \citep[AutoML; see][for an overview]{hutter-book19a} has alleviated practitioners from this manual task and support users in achieving peak performance of machine learning (ML) systems.
However, since the selection of these ML design decisions is automatically made by the optimization process, the transparency decreases, and the question of how and why a particular pipeline or configuration was chosen remains open. The lack of insights in current AutoML systems~\citep{drozdal-iui20} goes so far that some users even prefer manual tuning as they believe they can learn more from this process~\citep{hasebrock-arxiv22a}. Even more importantly, in safety-critical applications like automated driving or the medical domain, transparency and interpretability of both the ML model and the AutoML process are strongly needed to gain trust in the systems.

The brisk development of machine learning \citep{pedregosa-online22} and AutoML~\cite[e.g.][for neural architecture search]{lindauer-jmlr20} creates new challenges for interpretation and analysis tools. For example, \emph{multi-objective} optimization \citep[see, e.g., ][]{mietten-book98} allows for simultaneously optimizing all possible non-dominated trade-offs of multiple objectives (e.g., minimizing network size while maximizing classification accuracy \citep{benmeziane-arxiv21a}). Further, incorporating multiple budgets into the process (a.k.a. \emph{multi-fidelity} optimization) is used to speed-up optimization by estimating the final performance on lower budgets such as epochs~\citep{falkner-icml18a,li-jmlr18a} or subsets of the data~\citep{jamieson-aistats16a,klein-aistats17}. Given these multitudes of different approaches and considering that optimization runs may still take several days, depending on the problem, interpretation tools for AutoML need to be flexible, interactive, and ad-hoc.

We introduce \dc{}, successor of \cave\footnote{CAVE stands for Configuration, Evaluation, Visualization and Evaluation}~\citep{biedenkapp-lion18a}, an interactive dashboard for analyzing AutoML runs to bring the human back into the loop.
Our framework mainly focuses on Hyperparameter Optimization~\citep[HPO]{bischl-arxiv21a} but can also be used with Neural Architecture Search~\citep[NAS]{elsken-jmlr19} and Combined Algorithm and Hyperparameter Selection~\citep[CASH]{thornton-kdd13a} by encoding architectural, algorithmic choices or ML pipelines as part of the configuration.
\dc{} enables users to interactively explore and analyze outputs of optimizers in the form of texts, tables, and graphics already \emph{while} the optimization process is running. Users can dive \emph{deep}er into their particular interesting topics as our proposed framework is structured in modules (or plugins), all of which complement each other to get an idea of what happened behind the black-box facade of the optimizer. 

\dc{} is designed with AutoML for deep learning in mind, a.k.a. AutoDL. 
First, the real-time monitoring of \dc{} allows users to quickly gain insights even for costly and long-running optimization runs, as a typical trait of AutoDL.
Second, pure black-box optimization is too expensive for training several DNNs; hence, multi-fidelity optimization is one of the key approaches to making AutoDL feasible~\citep{zimmer-ieee2021a}. \dc{} is the first tool to support analysis of multi-fidelity optimization procedures.
Last, with the diverse applications of deep learning, several objectives need to be taken into account by AutoDL.
Therefore, we believe that \dc{} is well suited to become a go-to tool for deep learning practitioners using AutoML.

In this paper, we explain how certain framework design decisions benefit the users and why \dc{} can be easily used by different user groups, including researchers, data scientists, and machine learning engineers. Finally, we show how our framework can be utilized to answer crucial questions:
\begin{inparaenum}[]
    \item Have there been any issues in the optimization process?
    \item How much of the space is already covered?
    \item Are the selected budget steps efficient?
    \item Which configuration should be selected if multiple objectives are important?
    \item Which hyperparameters and design decisions are the most important ones?
\end{inparaenum}

Our contributions are twofold:
\begin{inparaenum}[i)]
    \item We present a novel framework that enables analysis of AutoML optimization processes while increasing the transparency and bringing the human back in the loop.
    \item In an exemplary study on outlier detection, we show how our framework can be used to answer questions typically faced when using AutoML tools.
\end{inparaenum}

\section{Framework}

We begin by giving an overview of \dc{} and describe how a typical workflow looks like. Finally, we explain how certain requirements shaped the implementation.

\paragraph{Overview}

\begin{figure}[tb]
    \centering
    \includegraphics[width=0.9\textwidth]{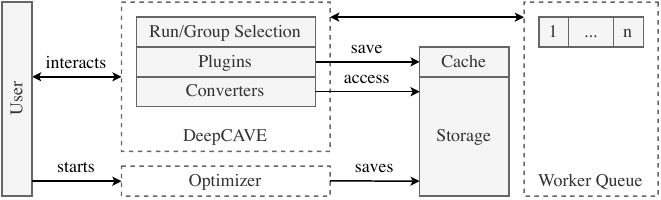}
    \caption{\centering Relation between user, optimizer and \dc.}
    \label{fig:overview}
\end{figure}

The interconnection between a user, optimizer, and \dc{} is shown in Figure~\ref{fig:overview}.
After starting \dc, the user can select optimizer runs or stagger runs as groups.
\dc{} accesses the optimizer data on the file system via \emph{converters}, which monitor both finished processes and running processes that regularly write new results to disk, as long as a suitable converter is available.
At release, \dc{} natively supports well-known AutoML packages such as BOHB~\citep{falkner-icml18a}, DEHB~\citep{awad-ijcai21a}, SMAC~\citep{lindauer-jmlr22a}, Auto-Sklearn~\citep{feurer-jmlr22a}, and Auto-PyTorch~\citep{zimmer-ieee2021a}, but also provides a Python interface to record the optimizer's output directly.
After converting the data, the user can start interpreting the optimization process through the plugins, all of which can incorporate texts, tables, and interactive graphics to show useful information such as simple statistics, hyperparameter importance, or how an optimization run behaves over time.
Since some plugins might require more heavy computation, a worker queue ensures responsiveness at all times.

\paragraph{Design Decisions}

\dc{} is implemented on top of Dash~\citep{plotly-url15} and supports interactive visualizations natively provided by Plotly.
With the ease of extensibility in mind, our framework is written purely in Python as it is the most common programming language in the field of ML~\citep{raschka-info20}.
Thus, (Auto)ML researchers have a low barrier of entry when designing new plugins for \dc{}.
Moreover, we structured plugins into input, filter, and output blocks, all of which are rendered by easy-to-understand HTML-like elements. All plugins have access to various base functionalities due to a unified run interface, which is instantiated based on the selected runs or groups by a converter at runtime. 
We integrated a powerful interface to provide maximal flexibility to all plugins: The run instance holds meta data, configurations $\bm{\lambda} \in \Lambda$ from the configuration space $\Lambda$, objectives $C = \{c_1, \ldots, c_n\}$, budgets $B = \{b_1, \ldots, b_m\}$ and further helper methods. Additionally, the whole optimization history with its $K$ trials $\{\lambda^k, b^k,C(\lambda^k, b^k), \mathrm{status}, \mathrm{additional}\}_k^K$ is saved, in which each trial consist of a configuration $\bm{\lambda}^k \in \Lambda$, a budget $b^k \in B$ , and the obtained objective values $C(\lambda^k, b^k)$. Furthermore, trials are marked with a status (e.g., successful, crashed or timed-out) and equipped with additional information (e.g., traceback). 

\dc{'s} dynamically generated texts, hover tooltips, and integrated documentation enables and assist users in understanding the plugins and interpreting the data efficiently. Finally, the integrated cache ensures the quality of life improvements as repeated requests do not trigger redundant computation with the current state of the run in mind. In particular, hashes of the optimizer's outputs are compared, deciding whether both the cache and run instance are up-to-date or have to be updated.

\section{Exemplary Study using \dc}

\begin{figure}[tb]
    \centering
    \subfloat[\centering Trial Statuses]{\includegraphics[width=0.5\textwidth]{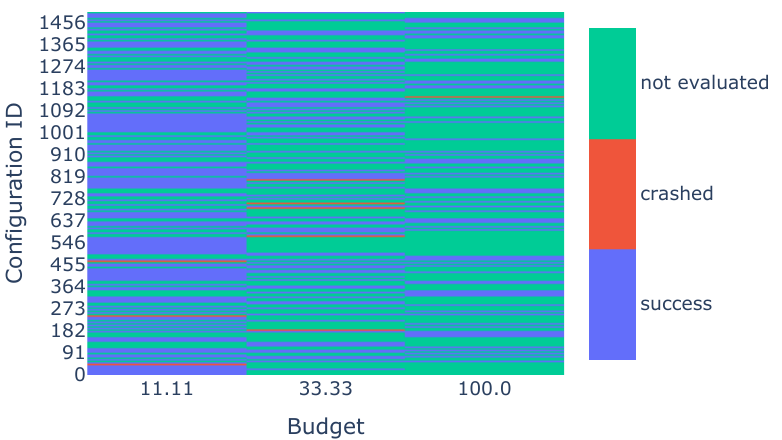}\label{fig:statuses}}
    \subfloat[\centering Configuration Footprint]{\includegraphics[width=0.5\textwidth]{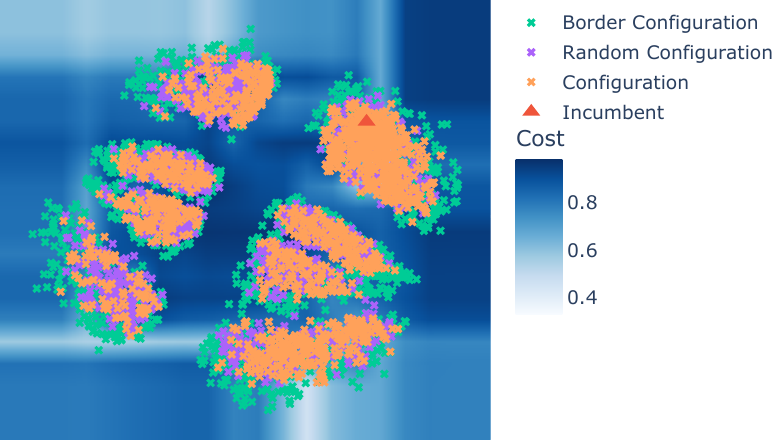}\label{fig:cf}}\hfill
    \subfloat[\centering Pareto Front
    ]{\includegraphics[width=0.5\textwidth]{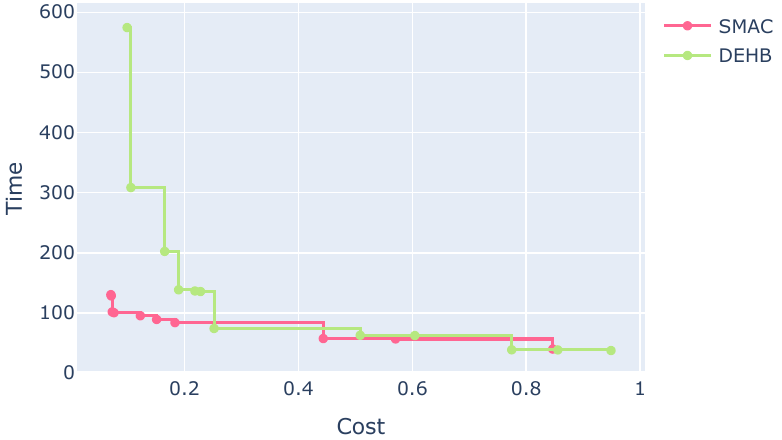}\label{fig:pf}}
    \subfloat[\centering Hyperparameter Importance]{\includegraphics[width=0.5\textwidth]{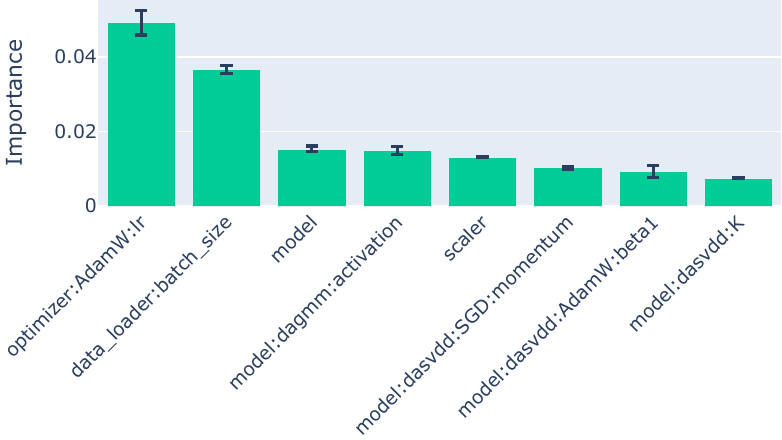}\label{fig:hi}}
    \caption{\centering\dc{} plots using a group of three SMAC runs on the highest budget.}
    \label{fig:experiments:hyperparameter_analysis}
\end{figure}

In the following, we demonstrate \dc{'s} capabilities and answer questions that typically arise when using AutoML tools.
We emphasize that \dc{} is a modular system that can potentially answer many more questions than we explore in this example study. We provide example questions for each plugin in the documentation (\url{https://automl.github.io/DeepCAVE}) to help the user to choose the right plugin for each question.

\paragraph{Experimental Setup}
Our exemplary ML-task is outlier-detection on the \emph{pendigits}~\citep{keller-ieee12} dataset with a contamination ratio of 15\%. We use two HPO optimizers SMAC3~\citep{lindauer-jmlr22a} and DEHB~\citep{awad-ijcai21a}, both of which maximize the area under the precision-recall (AUPR) curve to determine a well-performing DL pipeline. Since the optimizers minimize by default, we refer to \emph{cost} as $1-\mathrm{AUPR}$ in the following. The configuration space consists of 39 hyperparameters, in which the optimizer has to select a model between AE~\citep{rumelhart-book85}, VAE~\citep{kingma-arxiv13a}, DASVDD~\citep{hojjati-corr21}, and DAGMM~\citep{zong-iclr18a}. The optimizers make use of epochs for multi-fidelity optimization with budgets $\{11, 33, 100\}$.

\paragraph{Q1: Have there been any issues in the optimization process?}

The question can be answered using the \textit{Overview} plugin. In addition to displaying meta data, the optimizer's budgets and objectives, \dc{} dynamically generates a short status report:
\begin{quote}\vskip -.125in
    Taking all evaluated trials into account, 96.66\% have been successful. The other trials are crashed (3.24\%). Moreover, 47.96\%/30.11\%/21.78\% of the configurations were evaluated on budget 11.11/33.33/100.0, respectively.
\end{quote}\vskip -.05in

The status report is further supported by a barplot and a heatmap and is ideal for sanity checks. The heatmap (see Figure~\ref{fig:statuses}) shows the status of trials. In particular, crashed configurations are not evaluated on higher budgets, and only some configurations are performing well enough to invest more computational resources. The plugin offers further details to understand why certain trials have failed. Unsuccessful configurations are listed with their status and traceback, making it straightforward to identify the issue.

\paragraph{Q2: How much of the space is already covered?}
The plugin \textit{Configuration Footprint} reveals the structure and coverage of the configuration space. Using the multi-dimensional scaling (MDS), it generates a 2D projection of the configuration space~\citep{biedenkapp-lion18a,zoller-arxiv22a}, see Figure~\ref{fig:cf}. Both evaluated (orange) and unevaluated configurations are shown in the plot, with the latter further divided into border\footnote{A border configuration uses only hyperparameters with min and/or max bounds.} (green) and random (purple) configurations.
Border configurations ensure that the entire space is spanned, whereas the random configurations highlight unexplored areas.
The plot thus helps users to see if the optimizer already densely sampled the space or might need more evaluations to achieve better coverage.
We note that, in this example, not all parts of the MDS space can be covered because of the hierarchical structure induced by the configuration space.

\paragraph{Q3: Are the selected budget steps efficient?}
For multi-fidelity optimization, it is desirable to use budgets such that the performance across budgets is strongly correlated since multi-fidelity allows an optimizer to discard poorly performing configurations on the lower budgets quickly. In the plugin \textit{Budget Correlation}, the correlation between all budget combinations is shown. \dc{} tells us directly that all of our budget combinations have a very strong correlation in our example. Hence the chosen budgets are appropriate -- not shown here.

\paragraph{Q4: Which configuration should be selected if both cost and time are important?}
\dc{} can highlight the best configurations with respect to both objectives using the plugin \textit{Pareto Front}. The lines in Figure~\ref{fig:pf} depict the Pareto front for cost and (training) time. Depending on the application, a user could easily choose a configuration optimized for time, cost, or somewhere in between. This plugin also supports comparison between multiple runs or groups so that, in our case, \textit{SMAC} and \textit{DEHB} can be quickly compared. Hovering over a point reveals the configuration, its objective values on the highest budget, and selected hyperparameters. The points are clickable, taking the user directly to a detailed page with information about its origin, objective values, hyperparameter visualizations, and auto-generated code that can be directly copied to use the configuration in Python.

\paragraph{Q5: Which hyperparameters are the most important ones?}
The plugin \textit{Importances} let us choose between LPI~\citep{biedenkapp-lion18a} and fANOVA~\citep{hutter-icml14a}. 
The plugin's output (see fANOVA results in Figure~\ref{fig:hi}) shows that the learning rate, batch size, and model have the largest impact on performance. However, the plot does not reveal, e.g., which model was used in the best-performing configurations. For this, we can use the \textit{Configuration Footprint} (Figure~\ref{fig:cf}) again. Having a closer look at the incumbent, i.e., the best performing configuration, and using hover information from its neighbors, we find that \textit{DAGMM} is the best choice on the given dataset. In a further optimization run, we could utilize this information, e.g., to prune the configuration space, focusing only on more promising regions that include the incumbent's neighborhood.

\section{Related Work}

The design of \dc{} is driven by the need to make AutoML more trustworthy by means of increasing transparency. Other tools, such as CAVE~\citep{biedenkapp-lion18a} for algorithm configuration, IOHanalyzer~\citep{doerr-asc20a} for discrete optimization and XAutoML~\citep{zoller-arxiv22a} for AutoML, strive for similar goals, but have major limitations compared to \dc.
First of all, all three can only be used \textit{after} the optimization process has finished.
This restriction prevents users from discovering potentially non-fatal errors while the optimizer is running.
Furthermore, \dc{} is the only tool that is designed with multi-objective and multi-fidelity optimization in mind.

\cave{} only generates static reports, which have several drawbacks, e.g., it is nigh on impossible to explore a specific problem deeper.
Since \cave~is designed for experts only, it is challenging for inexperienced users to understand the given information.
Our novel framework mitigates these shortcomings by providing an interactive interface, intuitive visualization, and online documentation directly available where users need it, i.e., when making choices and when results are presented.
Thus, we view \dc{} as a tool that is both accessible and useful for AutoML users of all experience levels.

XAutoML focuses on analysis for classification tasks and does not support multi-fidelity or multi-objective optimization. \dc{} mitigates these shortcomings by being optimizer agnostic, enabling the use of multi-fidelity and multi-objective optimization, thereby avoiding a strong focus on specific target applications. 
Although \dc{} is built for AutoML, in principle, it can be used to analyze any black-box function optimizer.

\section{Conclusion}

We introduced \dc{}, an interactive framework to analyze AutoML runs in real-time, and discussed in detail how our design decisions make our framework easy to use with \emph{any} multi-fidelity AutoML optimizer as well as its extensibility through design. Based on an exemplary study of outlier detection, we demonstrated how central questions commonly faced by users of AutoML tools could be answered. This study further shows how our tool improves the transparency of AutoML and how the application aids the user's understanding of the optimization process in-depth. Finally, we believe that our interactive tool \dc{} and its diverse collection of default plugins bring us closer to human-centered AI and will help to increase the trustworthiness of AutoML tools.

\section*{Acknowledgement}
René Sass and Marius Lindauer acknowledge financial support by the Federal Ministry for Economic Affairs and Energy of Germany in the project CoyPu under Grant No. 01MK21007L. Eddie Bergman acknowledges support by TAILOR, a project funded by EU Horizon 2020 research and innovation programme under GA No 95221.
Frank Hutter acknowledges funding by European Research Council (ERC) Consolidator Grant ``Deep Learning 2.0'' (grant no.\ 101045765). Funded by the European Union. Views and opinions expressed are however those of the author(s) only and do not necessarily reflect those of the European Union or the ERC. Neither the European Union nor the ERC can be held responsible for them.
\begin{center}\includegraphics[width=0.3\textwidth]{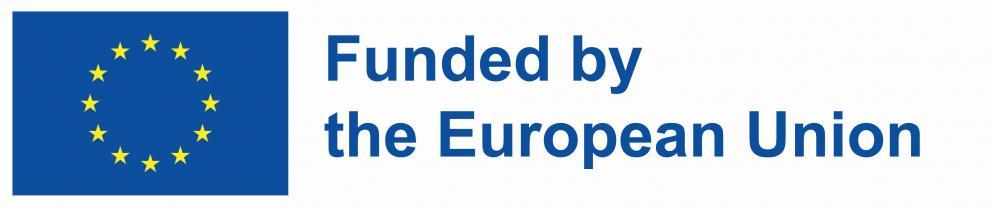}\end{center}

\bibliography{bib/strings,bib/lib,bib/local,bib/proc}

\end{document}